\documentclass[conference]{IEEEtran}

\let\labelindent\relax

\usepackage{times}

\usepackage[numbers,sort,compress]{natbib}
\usepackage{multicol}
\usepackage[usenames,dvipsnames,table]{xcolor}
\usepackage[bookmarks=true]{hyperref}
\hypersetup{
    colorlinks=true,
    linkcolor=MidnightBlue,
    filecolor=MidnightBlue,      
    urlcolor=MidnightBlue,
    citecolor=MidnightBlue,
}
\usepackage{pdfx}

\usepackage{amsmath}
\usepackage{amssymb}
\usepackage{mathtools}
\usepackage{amsthm}
\usepackage{enumitem}
\usepackage{cleveref}
\usepackage{caption}
\usepackage{soul}
\usepackage{booktabs}
\usepackage{subcaption} 

\newcommand{\myitem}{\vspace{1mm}\item}

\newcommand{\ignore}[1]{}

\newcommand{\ours}{UVA}

\usepackage[dvipsnames]{xcolor}
\definecolor{rowblue}{RGB}{220,230,240}
\definecolor{myorchid}{RGB}{150,10,30}
\definecolor{myblue}{RGB}{10,30,250}
\definecolor{mygreen}{RGB}{10,120,10}

\usepackage{amsmath,amsfonts,bm}

\def\eqref#1{equation~\ref{#1}}

\def\1{\bm{1}}

\DeclareMathAlphabet{\mathsfit}{\encodingdefault}{\sfdefault}{m}{sl}
\SetMathAlphabet{\mathsfit}{bold}{\encodingdefault}{\sfdefault}{bx}{n}

\begin{document}

\title{Unified Video Action Model}

\author{
\authorblockN{
\textbf{Shuang Li} \quad
\textbf{Yihuai Gao} \quad
\textbf{Dorsa Sadigh} \quad
\textbf{Shuran Song}}
\authorblockA{Stanford University}
\vspace{2mm}
\authorblockA{
\textbf{\textcolor{magenta}{\url{https://unified-video-action-model.github.io/}}}
}
}


%


\twocolumn[{%
\renewcommand\twocolumn[1][]{#1}%
\maketitle
\includegraphics[width=1\linewidth]{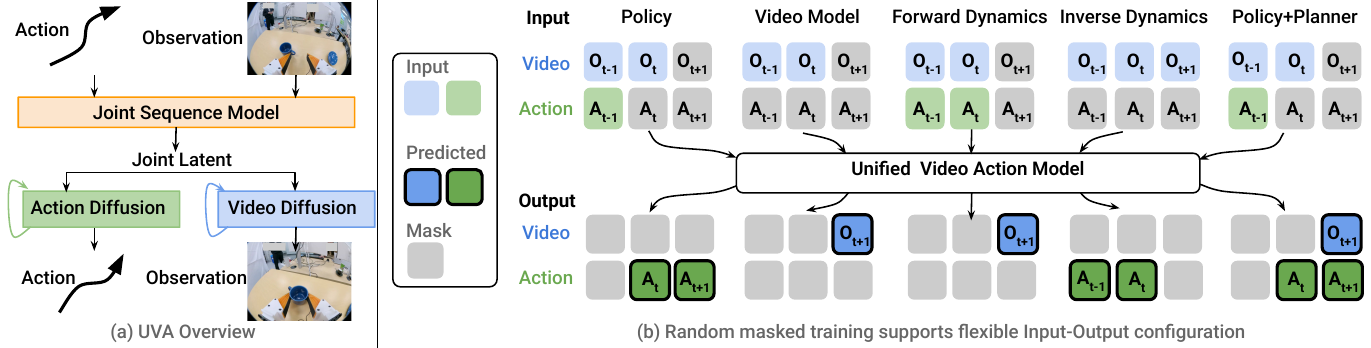}
\captionof{figure}{\small \textbf{Unified Video Action Model.} (a)
UVA features a \textit{joint} video-action latent representation and \textit{decoupled} video-action decoding.  The joint latent representation effectively models the underlying relationships between video and action sequences, while the decoupled diffusion enables high-speed action inference by bypassing video generation.
(b) By leveraging masked training, \ours~supports flexible input-output combinations for actions and videos. This versatility allows the model to function as a robot policy, video model, forward or inverse dynamics model, or even a combined policy and video planner—all within a unified framework.}


\label{fig:overview}
\vspace{2mm}
}]


\begin{abstract}

A unified video and action model holds significant promise for robotics, where videos provide rich scene information for action prediction, and actions provide dynamics information for video prediction. However, effectively combining video generation and action prediction remains challenging, and current video generation-based methods struggle to match the performance of direct policy learning in action accuracy and inference speed.
To bridge this gap, we introduce the Unified Video Action model (UVA), which jointly optimizes video and action predictions to achieve both high accuracy and efficient action inference. The key lies in learning a joint video-action latent representation and decoupling video-action decoding. The joint latent representation bridges the visual and action domains, effectively modeling the relationship between video and action sequences. Meanwhile, the decoupled decoding, powered by two lightweight diffusion heads, enables high-speed action inference by bypassing video generation during inference. Such a unified framework further enables versatile functionality through masked input training. By selectively masking actions or videos, a single model can tackle diverse functions beyond policy learning, such as forward and inverse dynamics modeling and video generation. Via an extensive set of experiments, we demonstrate that UVA can serve as a general-purpose solution for a wide range of robotics tasks without compromising performance compared to methods tailored for specific applications. 
Results are best viewed on \href{https://unified-video-action-model.github.io/}{our website}. 
\end{abstract}
\section{Introduction}
\label{sec:introduction}




A unified video and action model that jointly learns an agent's actions and their effects on visual observations holds great promise for robotics -- videos provide rich environmental context for predicting actions, while actions reveal how interactions drive visual changes, enabling more accurate modeling of real-world dynamics.
However, despite its promise, previous approaches have often failed to fully realize this potential.
A key challenge lies in the inherent mismatch between the requirements of action and video generation. Action modeling demands \textbf{high temporal speed} to capture dense, fine-grained motions, while video generation requires \textbf{high spatial resolution} to produce high-fidelity visual outputs, which often results in slower processing speeds.  


Previous policy learning approaches have struggled to balance these conflicting requirements, often focusing on one aspect at the expense of the other. For instance, \textit{action only} methods like \cite{chi2023diffusion,zhao2023learning,kim2024openvla} bypass video generation entirely. While such approaches reduce computational complexity, they overlook the benefits of video generation -- adding observation supervision helps the model learn scene dynamics, which reduces overfitting to action history and enhances robustness to visual disturbances.
On the other hand, \textit{video generation} methods such as \cite{du2024learning,liang2024dreamitate} often first generate high-resolution videos and then predict actions based on the generated videos. While this hierarchical approach can utilize existing video models, it also introduces significant drawbacks, including slower processing speeds and the propagation of errors from the generated video into action prediction.


To address these limitations, we propose \textbf{\ours}, a \textbf{Unified Video and Action} Model designed to simultaneously model videos and actions -- capturing the underlying interactions between visuals and actions to enhance task understanding, while maintaining high-speed action prediction during inference. 
We propose the following three design choices to achieve this:


\textbf{1) Unified Latent Video-Action Representation:}
\ours\ introduces a unified latent representation that integrates both visual and action data.
Unlike traditional \textit{video generation} based policy methods which rely on a hierarchical video and action generation, \ours\ is trained simultaneously with supervision from both video and action data. This enables the model to capture the intricate dynamics shared between the visual and action domains with reduced computational overhead. Utilizing the rich scene information encoded in the latent representation unlocks \ours's superior performance in understanding complex environments and delivering precise action predictions.

\textbf{2) Decoupled Video-Action Diffusion for Fast Inference:}
To further enhance efficiency and achieve inference speed comparable to \textit{action-only} methods, \ours\ decouples video generation from action prediction. During training, the model employs two lightweight diffusion heads to decode video observations and actions from the unified latent space. At inference, this decoupling allows the system to bypass video generation entirely, directly utilizing the latent representation for fast action prediction. 
This design enables real-time policy deployment without sacrificing performance, as it still retains the rich representations learned during training from both visual motions and robot action trajectories.



\textbf{3) Mask Training for Flexibility:} The ability to predict both videos and actions through unified representations further unlocks the potential to perform a diverse set of functions using masked training. \ours\ can handle versatile functions that go beyond traditional policy learning by masking inputs and outputs as needed, as illustrated in \Cref{fig:overview}. 
This versatility enables the model to tackle complex scenarios, such as operating as a forward or inverse dynamics model, learning effectively from video-only datasets where action labels are unavailable, or simultaneously performing both low-level control and high-level planning.

We evaluate \ours~on seven publicly available benchmarks to assess its diverse capabilities. \ours~outperforms or matches state-of-the-art approaches, demonstrating particularly strong performance in multi-task settings. 
For instance, \ours~outperforms the best baseline by 20\% in success rate on PushT Multitask~\cite{chi2023diffusion, florence2021implicit} and by 5\% on Libero10~\cite{liu2024libero}. 
The experiments show that \ours~can serve as a general-purpose framework for different robotics tasks without compromising performance compared to methods tailored for specific applications. In sum, the unified video action model is:
\begin{itemize}
    \myitem  \textbf{Capable:} \ours~matches the state-of-the-art approaches that are tailored for robot policy learning \cite{chi2023diffusion,kim2024openvla} or planning \cite{du2024learning}, especially for multi-task learning.

    \myitem  \textbf{Practical:} The use of decoupled diffusion heads eliminates the need for video generation during policy inference, and the use of lightweight diffusion heads reduces the costs of the denoising process.     
    As a result, \ours~achieves a similar speed as Diffusion Policy \cite{chi2023diffusion}, making it practical for robot applications.

    \myitem  \textbf{Versatile:} Beyond policy learning, \ours~can also serve as a forward dynamics model for planning, as an inverse dynamics model to generate actions, a video generation model, or a combined policy and video planner.
\end{itemize}

\section{Related Work}
\label{sec:related_work}

There is a rich literature on video generation where the dominant approach includes diffusion-based methods \cite{sohl2015deep,ho2020denoising,song2020score,blattmann2023stable, ho2022video, ho2022imagen, brooks2024video, girdhar2023emu,song2023consistency,song2023improved}
or autoregressive-based methods \cite{weissenborn2019scaling, yan2021videogpt, villegas2022phenaki,deng2024autoregressive,gao2024vid,weng2024art}. 
Our model utilizes the ideas from both of these techniques, as we will discuss in \cref{sec:method}. 
Another line of related work involves the use of masked training in robot learning.
In this section, we will review both video generation and masked training approaches used in robotics.

\smallskip
\noindent \textbf{Video Generation for Policy Learning:} Video models aid policy learning by simulating task dynamics and predicting future states. Models like \cite{du2024learning, liang2024dreamitate} leverage video generation techniques to produce high-quality videos, which are then used for action prediction. 
PAD \cite{guo2024prediction} jointly trains video generation and action prediction; however, it cannot predict future actions independently of future image generation, resulting in slower inference.
The work by \cite{xu2024flow} leverages video models to generate object flow as an intermediate representation, which captures physical interactions and is used to predict actions for skill transfer across different robotic embodiments and environments. In \cite{hu2024video}, a video diffusion model is fine-tuned on robotics tasks, with the latent representations from the predicted videos serving as inputs to a policy network for action prediction.
Existing approaches that utilize video generation for policy learning often suffer from slow inference speeds, making them impractical for real-world applications. Additionally, these methods often require auxiliary components, such as low-level policies \cite{du2024learning} or image-tracking techniques \cite{xu2024flow}, to extract actions from the generated videos. As a result, the final action accuracy suffers from compounded errors in video generation and action prediction.

\smallskip
\noindent \textbf{Video Generation as Dynamics Models:} Video models can serve as dynamics models by predicting future states conditioned on current observations and actions, enabling robots to simulate and plan tasks. GameGen-X \cite{che2024gamegen} introduces a diffusion transformer model for generating and controlling open-world game videos, enabling interactive simulations. Genie \cite{bruce2024genie} utilizes a foundation world model to transform static images into interactive 3D environments, providing rich simulations for embodied agents. Additionally, \cite{valevski2024diffusion} demonstrates how diffusion models can act as real-time game engines, generating dynamic and interactive scenarios to facilitate decision making. 
These advancements highlight the versatility of video generation models in robotic applications. In this work, we propose a unified video and action model, showcasing its ability to address both policy learning and dynamics modeling within a single framework.

\begin{figure*}
\begin{center}
\includegraphics[width=1\textwidth]{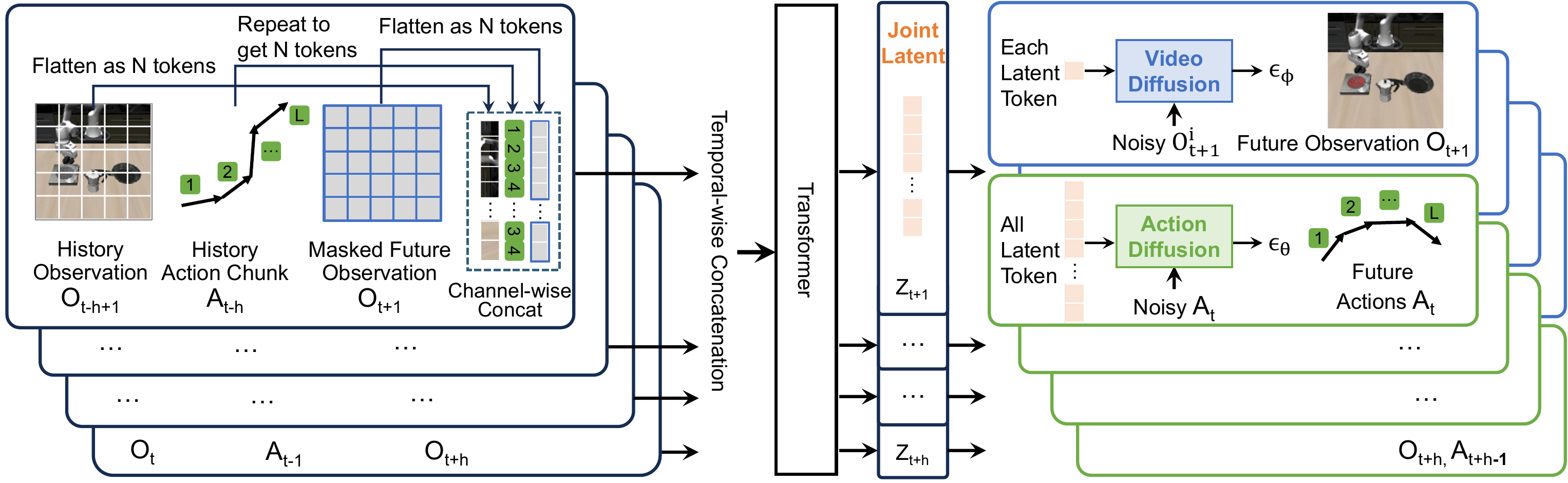}
\end{center}
\vspace{-5pt}
\caption{\small{\textbf{Network Architecture.} Given historical observations \(\{O_{t-h+1}, \dots, O_t\}\) and corresponding action chunks \(\{A_{t-h}, \dots, A_{t-1}\}\), the model predicts future observations \(\{O_{t+1}, \dots, O_{t+h}\}\) and actions \(\{A_t, \dots, A_{t+h-1}\}\). Each image observation is represented as a sequence of $N$ tokens, and each image corresponds to an action chunk with $L$ actions sampled at a higher frequency. During training, future observations are used with randomly masked tokens, while at inference time, the model starts from an empty image to predict the complete image. History observations, actions, and masked future observations are combined, passed through a Transformer, and decoded into actions and videos using diffusion heads.}}
\label{fig:model}
\vspace{-10pt}
\end{figure*}

\smallskip
\noindent \textbf{Masked Training:} 
Recent works in robotics have explored masked training techniques~\cite{radosavovic2023robot, liu2022masked, wu2023masked}. For example, Liu et al.~\cite{liu2022masked} and Wu et al.~\cite{wu2023masked} randomly mask observations and actions and reconstruct the missing portions. Their results show that masked training improves generalization to downstream tasks and enables the model to be used for various applications.
However, these methods primarily rely on low-dimensional state observations rather than videos, which are more natural but harder to predict. Radosavovic et al.~\cite{radosavovic2023robot} first pretrain a model to predict actions or observations using masked training and then finetune the model or use a linear probe for downstream tasks. Their work focuses solely on action prediction results. In contrast, our method does not require finetuning for downstream tasks and can be directly applied to various functions beyond policy learning, such as video generation, forward dynamics, and inverse dynamics.





\section{Unified Video Action Model}
\label{sec:method}


In robotics, we are interested in learning generalizable policies that map observations to actions. However, this objective often tends to overfit the training data, thereby limiting the ability of learned policies to adapt to new scenarios.
In contrast, video generation \cite{brooks2024video,veo2024} demonstrates strong generalization to novel scenes and supports training on datasets without actions. However, effectively leveraging video data for policy learning presents challenges such as the ability to match the high temporal speed required for outputting dense, fine-grained motions.
In this section, we discuss our approach to leveraging video-generation methods for robotics tasks.




\smallskip \noindent \textbf{Problem Statement:}
Given a sequence of image observations \(\{\mathbf{O}_{t-h+1}, \ldots, \mathbf{O}_{t}\}\) and action chunks \(\{\mathbf{A}_{t-h}, \ldots, \mathbf{A}_{t-1}\}\), where $h$ is the history horizon, our goal is to predict the future actions \(\{\mathbf{A}_{t}, \dots, \mathbf{A}_{t+h'-1}\}\) and observations \(\{\mathbf{O}_{t+1}, \dots, \mathbf{O}_{t+h'}\}\), where $h'$ is the future horizon. 
Each action chunk, e.g., $\mathbf{A_t} \in \mathbb{R}^{L\times m}$ consists of \(L\) actions, and each action has $m$ dimensions. 
We set $h=h'$ in the experiments. For simplicity, we refer to both as $h$ in the following sections.

We first introduce the model with complete video and action inputs and outputs (\S\ref{sec:method_encoder}-\S\ref{sec:decouple}). 
We then discuss how masked training can flexibly learn from any combination of video and action data (\S\ref{sec:method_mask_training}), enabling \ours~to perform various functions, including policy learning, video generation, forward and inverse dynamics, and integrated policy and planning.

As shown in \Cref{fig:model}, our method encodes the history of observations and actions (\S\ref{sec:method_encoder}), along with masked future observations (\S\ref{sec:mae}), and passes them to the Transformer \cite{vaswani2017attention}. 
For the masked observations, we randomly mask tokens within the future observation frames during training and train the model to reconstruct them. During inference, the model generates the full set of tokens, starting from an empty sequence.
In~\S\ref{sec:decouple}, we then discuss the choice of decoupling video-action diffusion for fast inference addressing the high temporal speed demand of robot policies.
\subsection{Encode History}
\label{sec:method_encoder}
We first process the historical image observations through a pre-trained VAE encoder (kl-f16) \cite{rombach2021highresolution} to obtain their latent representations. 
Each image is encoded into a latent map of dimensions \(\mathbb{R}^{w \times h \times c}\), where \(w\) and \(h\) represent the width and height, and \(c\) is the latent dimension. 
The map is then flattened and processed by a fully-connected (FC) layer, projecting each element into a \(d\)-dimensional latent vector. Thus, each image is represented as a sequence of $N$ visual tokens, each with \(d\)-dimensional features.

For history actions, we use a higher sampling frequency compared to observations, as observations typically exhibit redundancy and minimal changes over short time intervals. Each image observation (e.g., \(\mathbf{O}_{t-h+1}\)) corresponds to $L$ actions within an action chunk (e.g., \(\mathbf{A}_{t-h}\)).
We repeat the action chunk $M$ times to match the number of visual tokens as shown in \Cref{fig:model}. The repeated sequence is then passed through an FC layer, and converted into a sequence of $N$ action tokens, each with a \(d\)-dimensional latent representation.
These history visual and action tokens serve as conditions for predicting future observations and actions.

\subsection{Masked Autoencoder for Observation Prediction}
\label{sec:mae}
Our work is closely related to \cite{li2024autoregressive,chang2022maskgit}. Their method focuses on image generation conditioned on class labels. It begins by generating a subset of visual tokens for the image and then sequentially predicts additional tokens based on the previously generated ones, following an autoregressive process to complete the image. This step-by-step autoregressive approach has been shown to outperform the single-step generation of all visual tokens simultaneously.
To facilitate step-by-step prediction, they employ a masked autoencoder \cite{he2022masked} framework. During training, some visual tokens are randomly masked, and the model is trained to reconstruct these masked tokens. 

We follow this setting for video prediction.
Future observation frames \(\{\mathbf{O}_{t+1}, \ldots, \mathbf{O}_{t+h}\}\) are processed similarly to historical observations: they are passed through a VAE encoder to extract latent representations, followed by an FC layer, resulting in a sequence of $N$ tokens per frame, each with a \(d\)-dimensional latent vector. Some tokens are randomly masked out during training.
These visual tokens are concatenated channel-wise with historical visual tokens and action tokens, as shown in~\Cref{fig:model}, to form a new sequence of latent features.
Latents from $h$ different time steps are then temporally concatenated with latent representations from other time steps to produce a \(N \times h\) latent sequence.  
The resulting sequence is passed through a Transformer to fuse the video and action information, resulting in a set of joint video-action latent representations, \(\{\mathbf{Z}_{t+1}, \ldots, \mathbf{Z}_{t+h}\}\), where each latent (e.g., \(\mathbf{Z}_{t+1}\)) contain $N$ latent tokens. These joint video-action latent tokens are then used to reconstruct the future observations and corresponding action chunks.

For tasks that involve language instructions, such as Libero10~\cite{liu2024libero}, we incorporate language information using cross-attention in the Transformer. The language input is encoded into a \(d\)-dimensional token using the CLIP~\cite{radford2021learning} text encoder. To emphasize the language, this token is repeated \(M\) times and appended to the \(N \times h\) video-action tokens, resulting in a total of \(N \times h + M\) tokens. These tokens are then passed through the Transformer to fuse the multimodal information. We take the first \(N \times h\) outputs of the Transformer as the joint video-action latent representations, \(\{\mathbf{Z}_{t+1}, \ldots, \mathbf{Z}_{t+h}\}\).

To minimize information leakage across different frames, we consistently mask the same positions across all video frames. At inference time, the model generates complete videos by predicting all tokens starting from an empty sequence. 
At each autoregressive generation step, visual tokens at the same position across all video frames are generated simultaneously as shown in Supplementary \S\ref{apx:autogressive}.
Unlike image generation conditioned on class labels or text, historical observations provide rich contextual information about the environment. We found that a single-step generation is sufficient to generate high-quality observations, while using additional steps can further enhance the quality.



\subsection{Decoupled Video and Action Diffusions}
\label{sec:decouple}
Previous video generation-based policy learning methods rely on hierarchically generating videos first and then predicting actions, leading to slow speed and accumulated errors. To address this, we propose decoupling video and action prediction while training them jointly. 
During training, video generation helps the latent representations $\mathbf{Z}$ capture more detailed scene information, which benefits action prediction. 
During policy inference, where speed is crucial, the decoupled design allows us to skip video generation and decode only the actions. Similarly, for video generation, where quality is the priority, we can perform multi-step autoregressive video generation while bypassing action decoding.


We introduce two lightweight diffusion decoders for action and video prediction (see \Cref{fig:model}). 
Instead of performing the denoising over the entire model \cite{chi2023diffusion}, our approach restricts the denoising process to the lightweight decoders, delivering more efficient performance. This design preserves the generative strengths of diffusion models while significantly reducing inference time. 

The joint latent \(\mathbf{Z}\) serves as the conditioning input for the diffusion decoders. The video diffusion decoder processes each latent token \(z_i \in \mathbf{Z}_{t+1} = \{z_1, \ldots, z_{N}\}\) to predict individual patches in the video frame, which are then reshaped and sent to the VAE decoder to reconstruct the full frame $\mathbf{O}_{t+1}$. For the action diffusion decoder, all latent tokens in \(\mathbf{Z}_{t+1}\) are aggregated using a convolutional layer, followed by an MLP layer, to produce an action latent. This latent encodes both visual and action-related information for the current step and serves as the condition for the action diffusion model to generate the action chunk $\mathbf{A}_t$. We use the diffusion head (base size) from \cite{li2024autoregressive} for both action and video prediction.

During training, the decoders learn to predict the noise added to noisy action chunks or video patches. 
The action diffusion loss~\cite{sohl2015deep,ho2020denoising,song2020score} is defined as:
\[
\mathcal{L}_{\text{action}}(\mathbf{Z}, \mathbf{A}) = \mathbb{E}_{\epsilon, k} \left[\| \epsilon - \epsilon_\theta(\mathbf{A}^{(k)} | k, \mathbf{Z}) \|^2 \right],
\]
where \(\mathbf{A}^{(k)}\) represents the noisy actions, \(\epsilon\) is the added noise, \(k\) is the diffusion timestep, \(\mathbf{Z}\) is the joint video-action latent tokens, and \( \epsilon_\theta(\mathbf{A}^{(k)} | k, \mathbf{Z}) \) is the predicted noise.

Similarly, the video diffusion loss is defined as:
\[
\mathcal{L}_{\text{video}}(\mathbf{Z}, \mathbf{O}) = \mathbb{E}_{\epsilon, k} \left[\frac{1}{N} \sum_{i=1}^{N} \| \epsilon_i - \epsilon_\phi(\mathbf{O}^{i,(k)} | k, z_i) \|^2 \right],
\]
where \( \mathbf{O}^{i,(k)} \) represents the \( i \)-th noisy visual token in the video frame \( \mathbf{O}^{(k)} \) at diffusion timestep \( k \), \( N \) is the total number of visual tokens in a video frame, \( \epsilon_i \) is the added noise to the \( i \)-th visual token, $z_i$ is the latent token in $\mathbf{Z}$, and \( \epsilon_\phi(\mathbf{O}^{i,(k)} | k, z_i) \) is the predicted noise for the \( i \)-th token.


The total loss at each time step is the combination of the action and video diffusion losses:
$\mathcal{L} = \mathcal{L}_{\text{action}} + \mathcal{L}_{\text{video}}$. The overall loss is calculated as the sum of these losses over the time horizon $h$.
During policy inference or video generation, the decoders iteratively refine pure noise into actions or videos using the learned denoising process.

\subsection{Masked Training with Flexible Objectives} 
\label{sec:method_mask_training}
Instead of training the model solely on the task of predicting future observations and actions based on historical data, we propose a masked training approach with multiple training objectives using a unified framework. As illustrated in \Cref{fig:overview}, the model is trained on five distinct tasks by varying input and output combinations. Unused components are masked and replaced with a learned mask token. The action loss and video loss are selectively applied to supervise the model depending on the specific task. 

This training approach enables us to fully utilize the data in various combinations and supports the use of incomplete data, such as video data without corresponding actions.
This masked training strategy enables the model to perform a diverse range of functions, including acting as a robot policy, video model, forward and inverse dynamics model, and a combined policy and planner. For instance, when given only image observations, the model can function as an inverse dynamics model to generate action labels from videos. Additionally, this strategy helps prevent overfitting to specific tasks, enhancing the model’s overall versatility and robustness.



\section{Evaluation}
In the following sections, we evaluate \ours's capacities as policy \S\ref{sec:policy}, a video generator \S\ref{sec:eval_video}, a forward dynamics model \S\ref{sec:forward}, and finally an inverse dynamics model~\S\ref{sec:inverse}. 
In each scenario, we compare \ours~with methods that are specifically tailored for the corresponding application. 

\section{\ours~as Policy} 
\label{sec:policy}

We first investigate the effectiveness of \ours~on policy learning. 
As noted by Kim et al. \cite{kim2024openvla}, different policy designs excel in different settings. Their experiments show that while Diffusion Policy \cite{chi2023diffusion} performs better in single-task setups, it falls behind OpenVLA in multi-task scenarios.
To comprehensively evaluate the performance of \ours~as a policy, we conduct extensive evaluations in single-task and multi-task settings, and in simulated (\Cref{fig:env_fig_sim}) and real environments (\Cref{fig:env_fig_real}).
For simulation tasks, we used the same random seeds for different methods for a fair comparison.
\textbf{For real-world tasks, to minimize evaluation bias, all evaluations use public benchmarks with released datasets}—no additional training data were collected. 

\begin{table}[t]
    \centering
    \includegraphics[width=1\linewidth]{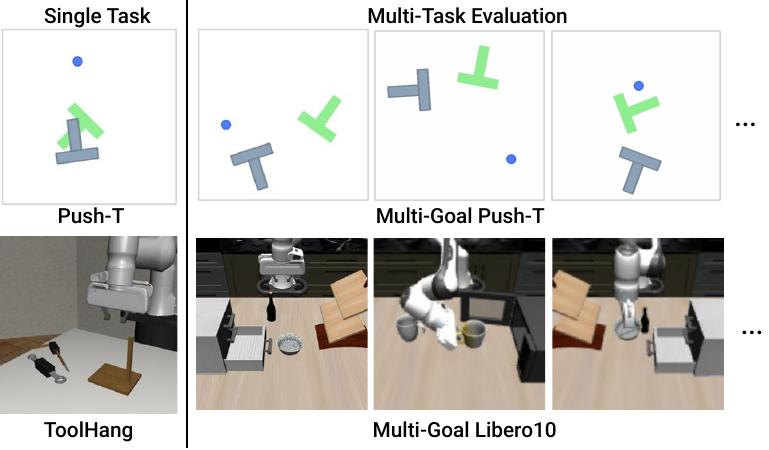}
    \captionsetup{type=figure}
    \vspace{-10pt}
    \caption{\small \textbf{Simulation Environments.} We evaluate \ours\ and baselines in both single-task and multi-task settings. In the multi-task scenario, the goal can be defined through the image observations (PushT-M) or language descriptions (Libero10).}
    \label{fig:env_fig_sim}
    %
\vspace{20pt}
\small
  \centering
  \setlength{\tabcolsep}{0.1em}
  \scalebox{0.93}{
  \begin{tabular}{l|cc|cc|r}
    \toprule
    & \multicolumn{2}{c|}{~~\bf Single-Task~$\uparrow$~~}  & \multicolumn{2}{c|}{~\bf Multi-Task~$\uparrow$~}  & {~\bf Speed} $\downarrow$  \\
    \midrule
    &  {~PushT} & {Tool~} & {~PushT-M~} & {~Libero10~} &  \\
    \midrule
    \bf DP-C~\cite{chi2023diffusion} & 0.91  & \textbf{0.95}  & 0.68 & 0.53 & 0.50s~~ \\
    \bf DP-T~\cite{chi2023diffusion} & 0.78  & 0.76 & 0.63 & 0.58 & 0.36s~~ \\
    \bf OpenVLA~\cite{kim2024openvla} & 0.35 & 0.18 & 0.22 & 0.54 & 1.52s~~ \\
    \bf UniPi~\cite{du2024learning} & 0.42 & 0.00 & 0.19 & 0.00 & 24.07s~~ \\
    \bf $\pi_0$~\cite{black2410pi0} & - & - & - & 0.85 & \bf 0.09s~~ \\
    \bf $\pi_0$-FAST~\cite{black2410pi0} & - & - & - & 0.60 & \bf 0.09s~~ \\
    \bf \ours-action & 0.45 & 0.62 & 0.46 & 0.86 & 0.22s~~ \\
    \bf \ours~ & \textbf{0.98}  & 0.88 & \bf 0.88 & \bf 0.90 & 0.23s~~ \\
    \bottomrule
  \end{tabular}
  }
  \captionsetup{type=table}
  \caption{\small \textbf{Policy Learning Results in Simulation.} UVA has higher success rate than the baselines in most settings, with a strong performance in multi-task scenarios. Speed is measured by a single action trajectory inference. All methods, except OpenVLA, infer 16 action steps per trajectory with 8 executed steps. OpenVLA infers one action at a time, so it is run 8 times to match the inference time for 8 executed actions.} 
  \label{exp:policy_results_sim}
\end{table}

\subsection{Simulation Benchmarks}
\label{sec:simulation_benchmark}
\vspace{2mm} \noindent
\textbf{Single-Task Evaluation:} 
We first evaluate single-task scenarios, where different policies are trained for different tasks. 
We compare \ours~with the baselines on the \ul{PushT}~\cite{chi2023diffusion, florence2021implicit} and \ul{Toolhang}~\cite{robomimic2021} tasks. We report the success rates of the best-performing checkpoint, averaging across 50 rollouts for \ul{PushT} and \ul{Toolhang}, respectively.



    

\vspace{1mm} \noindent\textbf{Multi-Task Evaluation:} 
We train one policy for multiple task goals defined by image or text.  We introduce a new task, \ul{PushT-M}, which extends the PushT task to include varying target ``T'' positions. We evaluate the best-performing checkpoint over 50 rollouts and report its average reward. \ul{Libero10}~\cite{liu2024libero} has 10 tasks. We evaluate each task in 50 different environments with varying random seeds and report the average rewards across all 10 tasks. 
See Supplementary \S\ref{apx:simulation_benchmark} for details. 



\subsection{Real-world Benchmarks}
\label{sec:realworld_benchmark}
\smallskip \noindent \textbf{Training Data:} We use two publicly available datasets introduced by \cite{chi2024universal} and \cite{lin2024data} without collecting any additional training data. Both benchmarks collect data using the handheld UMI \cite{chi2024universal} device. We used three tasks, including \ul{Cup Arrangement}, \ul{Towel Folding}, and \ul{Mouse Arrangement}, for training, and tested them on the ARX X5 arm.

\vspace{1mm} \noindent
\textbf{Single-Task Evaluation:} We train a single-task policy on the Cup task and directly compare it with the Diffusion Policy model provided by the author \cite{chi2024universal}, both trained on the same data. 
We evaluate each method over 20 rollouts with varying initial configurations and report the average success rate.


\begin{figure*}[t]
    \centering
    \includegraphics[width=\linewidth]{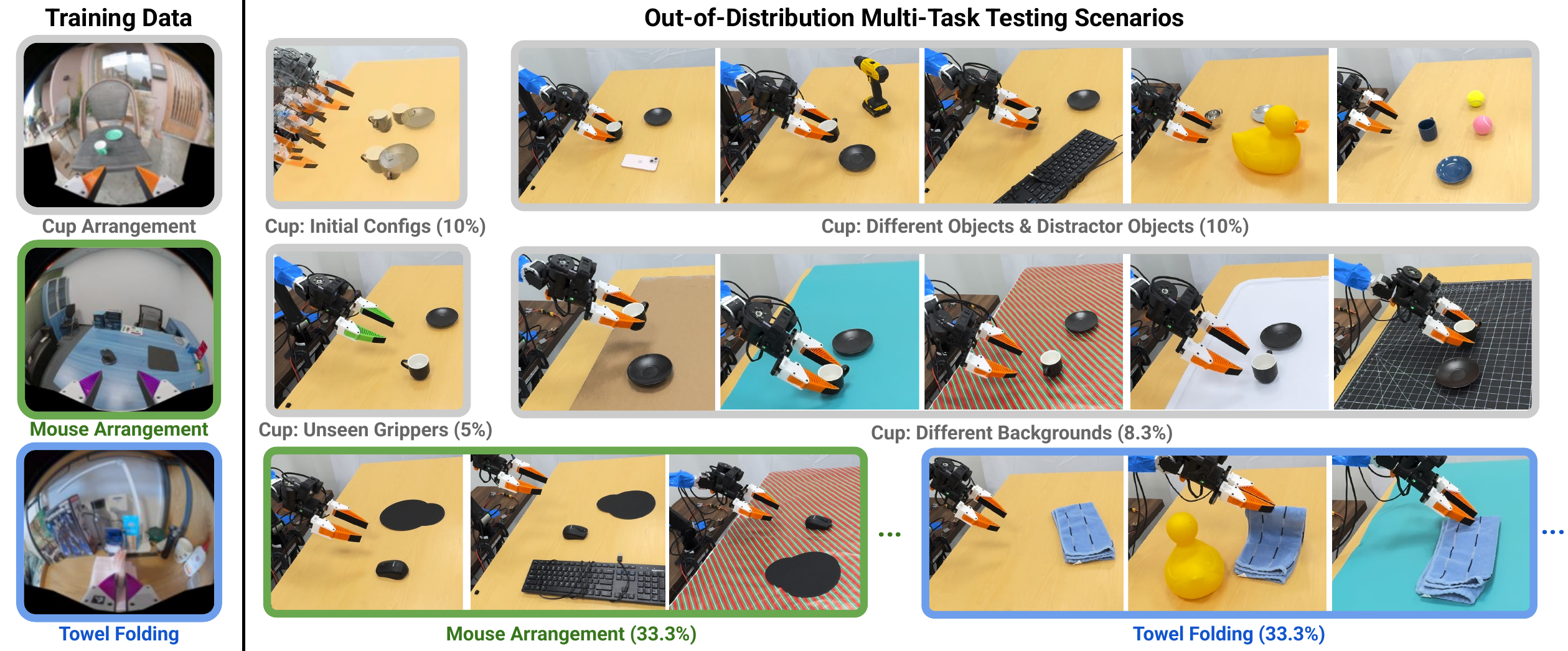}
    \caption{\small \textbf{Real-World Out-of-Distribution Evaluation.} We use the training data provided by prior works \cite{chi2024universal,weissenborn2019scaling}.  The test scenario is significantly out-of-distribution with unseen environments, objects, robots, and even gripper colors. The numbers in the parentheses show the percentage of such category of test cases in evaluation. Please refer to \href{https://unified-video-action-model.github.io/}{our website} for all evaluation rollouts. }
    \label{fig:env_fig_real}
    \small
    \vspace{-5pt}
\end{figure*}

\begin{table}[t]
  \centering
  \small
  \setlength{\tabcolsep}{0.5em}
\begin{tabular}{l|c|ccc|c}
    \toprule
    & \textbf{Single-Task $\uparrow$} & \multicolumn{3}{c|}{\bf OOD Multi-Task $\uparrow$} & \textbf{Speed~$\downarrow$} \\
    \midrule    
    & Cup & {~Cup}  & {Towel} & {Mouse~} & \\
    \midrule
    \bf DP-UMI \cite{chi2024universal} & \bf 0.95 & 0.50 & \textbf{0.70} & 0.40 & \bf 70ms \\
    \bf \ours   & 0.85 & \textbf{0.65}  & \textbf{0.70} & \textbf{0.80} & 95ms \\
    \bottomrule
\end{tabular}
  \captionsetup{type=table}
   \caption{\small \textbf{Success Rate on Real-World UMI \cite{chi2024universal} tasks.}  We compare \ours\ with DP-UMI which is designed for UMI tasks. \ours\ performs worse than DP-UMI in the single-task setting but achieves better performance in the multi-task setting. Both \ours\ and DP-UMI use 16 denoising steps. Speed is measured by inferring a single action trajectory consisting of 16 actions.
   }
   \label{exp:policy_results_real}
   \vspace{-7pt}
\end{table}


\vspace{1mm} \noindent
\textbf{Multi-Task Evaluation:} We train one model with all three tasks and then evaluate their performance on each task independently. We randomly selected 500 episodes from each dataset and combined them into a dataset to train both our model and Diffusion Policy \cite{chi2024universal}.
Since the training data were collected independently in prior works, all evaluation cases are \textbf{Out-of-Distribution} (OOD), involving unseen environments, objects, and robots. To ensure a wide testing distribution, we include cases with varying initial configurations, object distractors, background textures, and an unseen gripper color (green), as shown in \Cref{fig:env_fig_real}. Each policy is evaluated over 60 rollouts, with 20 rollouts per task.
See Supplementary \S\ref{apx:realworld_benchmark} for details.

    



\subsection{Baselines}
We compared with the following alternative methods, all methods are trained or fine-tuned on the same data as our model and tested using the same random seed and initial states. 

\begin{itemize}
    \item \textbf{Diffusion Policy} \cite{chi2023diffusion} is a state-of-the-art visuomotor policy model. We use both CNN-based network  [DP-C] and Transformer-based design [DP-T] from their original implementation for all simulation tasks. For real-world tasks, we used an improved Diffusion Policy \cite{chi2024universal}, which is optimized for UMI data. It leverages a CLIP-pretrained \cite{radford2021learning} ViT-B/16 \cite{alexey2020image} vision encoder, significantly improving visual understanding. We refer to it as [DP-UMI].

    \item \textbf{OpenVLA} \cite{kim2024openvla} is a state-of-the art vision-language-action (VLA) built on 7B Llama 2 \cite{touvron2023llama} for multi-task setting. It is trained on a diverse dataset encompassing a wide range of robots, tasks, and environments. We finetune OpenVLA on each task to optimize its performance.


    
    \item \textbf{$\boldsymbol{\pi_0}$} \cite{black2410pi0} is an open-source VLA model designed for general-purpose robot control. It employs a flow matching based architecture to generate continuous action sequences. 
    Trained on a diverse dataset spanning multiple robot embodiments and tasks, $\pi_0$ demonstrates strong zero-shot and fine-tuned performance. 
    We use the officially released checkpoints: $\pi_0$ and $\pi_0$-FAST, both finetuned on Libero tasks. 

    \item \textbf{UniPi} \cite{du2024learning} is a video-based policy model that generates videos first and then predicts actions based on the generated videos. Since the official implementation is not available, we used the code from \cite{ko2023learning}. This implementation relies on pixel-wise video generation, which results in slower video generation speed. For action inference, we train a model that processes two consecutive generated video frames using a pretrained ResNet-50 for the \textit{PushT} and \textit{PushT-M} tasks, and a ResNet-152 for the \textit{Tool Hang}, \textit{Libero10}, and \textit{Cup Arrangement}.

    \item \textbf{\ours-action} is an ablation of \ours, where the video generation part is excluded, and the model is trained solely as a policy model. This baseline aims to evaluate the effectiveness of joint video and action training.
\end{itemize}


\subsection{Policy Learning Results}
We evaluate policy learning results with UVA compared to the baseline methods on a few different axes: 1) action prediction accuracy, 2) inference speed, 3) robustness to visual disturbances, 4) robustness to history length, and 5) the effect of joint video-action modeling.

\smallskip \noindent \textbf{Action Prediction Accuracy (Simulation Tasks):} 
In \Cref{exp:policy_results_sim}, we compare \ours~with baseline methods in both single-task and multi-task settings. We use the same random seed for our method and baselines to ensure a fair comparison. 

\smallskip \noindent \ul{Simulation Single-Task}: Our method is able to match the performance of the state-of-the-art model DP-C and significantly outperform other video-based methods such as UniPi and vision-language-action model OpenVLA. 

\smallskip \noindent \ul{Simulation Multi-Task}: Our method is particularly strong in the multi-task setting.
Specifically, \ours~surpasses the best baseline by 20\% on the \textit{PushT-M} task and by 5\% on the \textit{Libero10} benchmark. 
This result demonstrates that \ours~model is able to better learn and leverage the general dynamics that are shared across tasks and, therefore, improve overall performance in the multi-task setting.   

Notably, $\pi_0$ (3.3B) and $\pi_0$-FAST (3.0B) are large models fine-tuned from pretrained checkpoints using the entire Libero dataset, while UVA is smaller (0.5B) and trained only on Libero10.
Moreover, $\pi_0$ and $\pi_0$-FAST use third-person and wrist-view images plus robot proprioception, while UVA relies only on third-person images. 
Despite its smaller size, limited input, and no external data, UVA outperforms $\pi_0$ and $\pi_0$-FAST.

\smallskip \noindent \textbf{Action Prediction Accuracy (Real-World Tasks):} 
\Cref{exp:policy_results_real} shows the results of real-world tasks. We ensure a fair comparison by keeping the initial placement of objects and grippers identical across different methods for each test rollout.

\smallskip \noindent \ul{Real-World Single-Task}: First, we evaluate the policy performance in a single-task setting. This evaluation aims to compare our method with a strong baseline in prior works by replicating a similar evaluation setup.  
Overall, \ours~performs comparable with DP-UMI, which is optimized with this particular training dataset. We noticed that the dataset contains extensive recovery data from the moments of failure to correct the policy. This data is particularly useful for models without history dependence, like DP-UMI, which can recover from the new observations included in the recovery data. In contrast, our model uses a longer history, which is advantageous for tasks requiring longer memory. However, in this case, the collected failure recovery data is less impactful for our model, as its longer memory window prioritizes learning from extended temporal patterns. While we could shorten our model’s history window, we maintain a consistent design across all tasks rather than tailoring it to this specific task and training data. 

\smallskip \noindent \ul{Real-World Multi-Task}: We train a single model using our method and evaluate it on three tasks individually. DP-UMI is trained and tested in the same manner for a fair comparison. For each task, we test the methods on 20 different cases, as shown in \Cref{fig:env_fig_real}. 
Our approach demonstrates superior performance in the multi-task setting, achieving a 15\% higher success rate on the Cup task and a 40\% higher success rate on the Mouse task compared to DP-UMI. 

In general, our method can successfully complete the task with distractor objects or changing backgrounds but fails when the background color is the same as the objects. Its visual understanding could be enhanced by training on additional video data without action labels.
DP-UMI performs well on the towel task but shows less stability when handling pick-and-place cups and grabbing the mouse. However, it performs well across different backgrounds due to its use of a pre-trained vision encoder.
When the gripper color is changed to an unseen green, the performance of both methods slightly decreases compared to the orange gripper used for training. However, both \ours~and DP-UMI show good generalization to the unseen gripper.
Please refer to \href{https://unified-video-action-model.github.io/}{our website} for details.

\begin{figure}[t]
    \centering
    \includegraphics[width=1\linewidth]{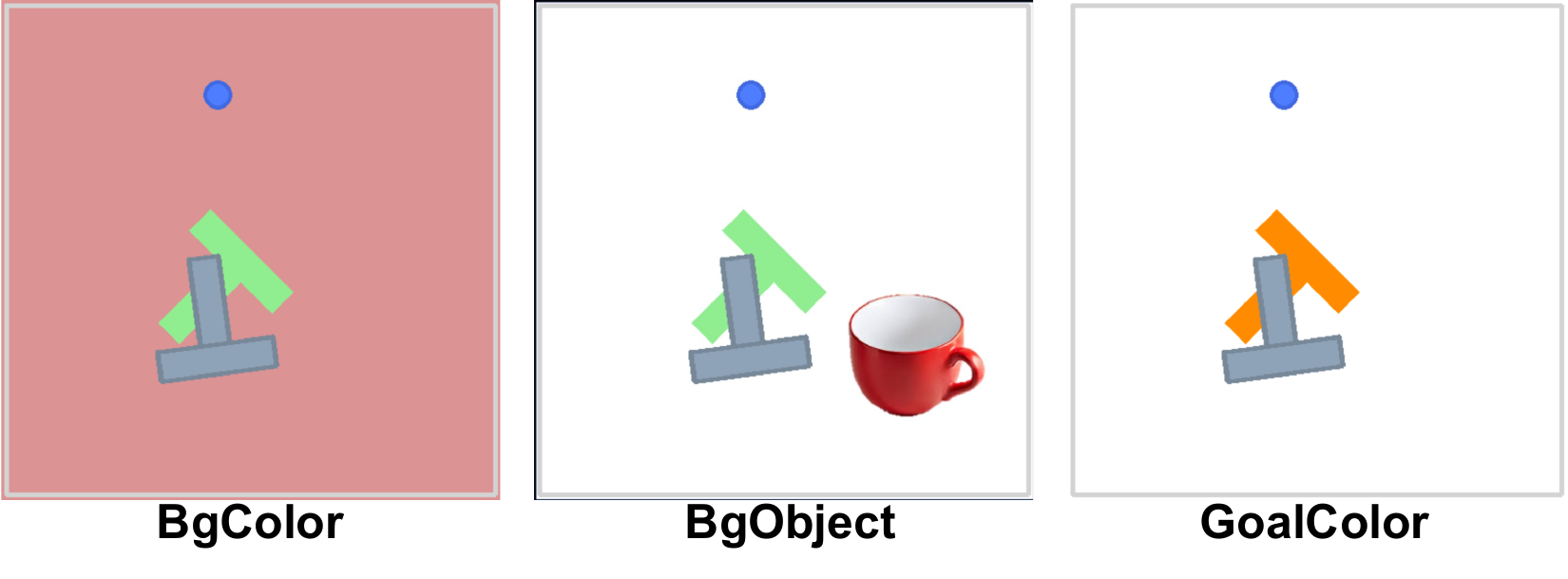}
    \vspace{-10pt}
    \caption{\small \textbf{Visual Disturbances on PushT.} Tasks are performed under altered visual conditions, including changes in background color, distracting background objects, and goal color.}
    \label{fig:pusht_figure}
    \vspace{15pt}
  \small
  
        \label{exp:vis_generalization}
  \centering
  \setlength{\tabcolsep}{0.2em}
  \scalebox{1}{
  \begin{tabular}{lcccc}
    \toprule
    & \bf Normal $\uparrow$ & \bf BgColor $\uparrow$ & \bf BgObject $\uparrow$ & \bf GoalColor $\uparrow$ \\
    \midrule
    \bf DP-C~\cite{chi2023diffusion} & 0.91 & 0.12 & 0.21 & 0.17 \\
    \bf DP-T~\cite{chi2023diffusion}  & 0.78 & 0.22 & 0.17 & 0.28 \\
    \bf OpenVLA~\cite{kim2024openvla}  & 0.35 & 0.17 & 0.13 & 0.32 \\
    \bf UniPi~\cite{du2024learning}  & 0.42 & 0.31 & \bf 0.36 & 0.40 \\
    \bf \ours~ & \bf 0.98 & \bf 0.35 & 0.31 & \bf 0.64 \\
    \bottomrule
  \end{tabular}
  }
  \captionsetup{type=table}
  \caption{\small 
         \textbf{Visual Generalization Results on PushT with Visual Disturbances.} Our method and UniPi, both video generation models, have higher success rates compared to other policy learning approaches.
         }
\end{figure}

\smallskip \noindent \textbf{Inference Speed:}  
Speed is evaluated based on a single action trajectory inference.
All methods, except OpenVLA, infer 16 action steps per trajectory with 8 executed steps. 
OpenVLA infers one action at a time, requiring 8 runs to match the inference time for 8 executed actions.
UniPi generates raw pixel videos, resulting in significantly slower inference.

Our method achieves faster inference speeds by performing diffusion iterations only on the lightweight action head, rather than the entire network as DP-C and DP-T. Thanks to the decoupled design, video generation can be skipped during policy inference, further improving efficiency. For simulation tasks, we use 100 denoise steps for action prediction. For DP-C and DP-T, we follow their original implementations and also perform denoising over 100 steps. With the same number of diffusion steps (\Cref{exp:policy_results_sim}), our method achieves faster inference compared to DP-C and DP-T.
Both $\pi_0$ and $\pi_0$-FAST achieve faster inference speed than other methods.

For real-world tasks (\Cref{exp:policy_results_real}), both \ours~and DP-UMI use 16 denoising steps for real-time manipulation. We found that the UVA Attention module in the Transformer accounts for half of the inference time, making \ours~slightly slower than DP-UMI. With future improvements, such as replacing the Attention with Flash Attention, our model could achieve faster speeds. 
We note that, although DP-UMI uses a pretrained ViT encoder, its model size (171.27M parameters) is smaller than DP-C (262.69M parameters). This explains why DP-C is slower than \ours~in \Cref{exp:policy_results_sim} and DP-UMI is faster in \Cref{exp:policy_results_real}.
Overall, \ours~achieves a good balance between speed and performance across diverse settings.

\smallskip \noindent \textbf{Robustness to Visual Disturbances:} 
We have shown that \ours~is robust to visual disturbances in the real-world multi-task setting in \Cref{fig:env_fig_real}. All tests are unseen during training, and even with more challenging distractor objects and backgrounds, \ours~ achieves higher success rates than DP-UMI.

To more rigorously evaluate this visual generalization capability, we perform a systematic evaluation in simulation by procedurally altering visual conditions in the \textit{PushT} environment. The modifications include changes to the background color, the addition of object distractor, and variations in goal color, as shown in \Cref{fig:pusht_figure}.
The evaluation was conducted on scenarios outside the training distribution, as the model was trained only in the standard environment in \Cref{fig:env_fig_sim}. The results show that video generation methods, such as UniPi and \ours, exhibit superior performance in handling visual disturbances. For example, with changes in goal color, UniPi achieves a success rate of 40\%, \ours~achieves 64\%, while OpenVLA only reaches 32\%.

\begin{figure}[t]
\begin{center}
\includegraphics[width=0.8\linewidth]{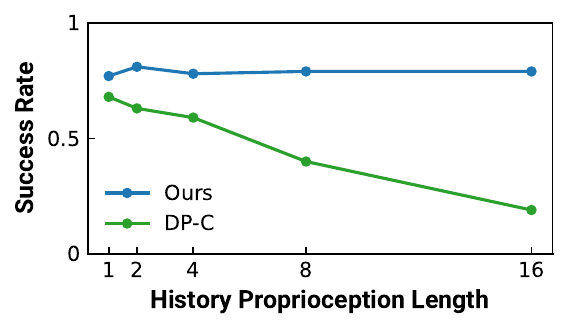}
\end{center}
\vspace{-10pt}
\caption{\small{\textbf{Robustness to History Length on PushT-M.} Typical policy learning frameworks such as DP-C \cite{chi2023diffusion} often experience performance drops with increased history length due to overfitting, while our model maintains robust performance. }}
\label{fig:difflen}
\vspace{-5pt}
\end{figure}

\smallskip \noindent \textbf{Robustness to History Length:} 
Prior policy learning methods, such as DP-C, often experience performance degradation as the history length increases as shown in \Cref{fig:difflen} evaluated on the \textit{PushT-M} task. 
In contrast, \ours~can effectively adapt to longer history inputs. 
By jointly predicting video and action, our model sustains robust performance even as the history length grows. This highlights the better potential of \ours~for tasks that require reasoning over extended temporal contexts.

\smallskip \noindent \textbf{Effect of Joint Video-Action Modeling:} We evaluate this by comparing \ours~with a baseline (\ours-action) that removes the video generation part. As shown in \Cref{exp:policy_results_sim}, this modification led to reduced performance compared to the complete framework, highlighting the critical role of video generation in improving policy learning.






\section{\ours~as a Video Generator}
\label{sec:eval_video}
\ours~can function as a video generation model by bypassing the action diffusion head during inference.
We compare \ours\ with UniPi on video generation results across two datasets: \textit{Libero10} and \textit{Cup Arrangement}, in \Cref{exp:video}. Performance is evaluated by computing the \textit{Fréchet Video Distance} (FVD) \cite{unterthiner2018towards} for 500 videos generated by each method. FVD is a metric for assessing video quality by evaluating visual fidelity and temporal coherence. It compares statistical properties of feature representations from real and generated videos, using a pre-trained Inflated 3D ConvNet \cite{carreira2017quo}. Lower FVD scores indicate greater similarity.

The masked autoencoder training in our method (\S\ref{sec:mae}) facilitates video generation in an autoregressive manner, where visual tokens are generated across all video frames in parallel during the first stage. In the subsequent stage, the next set of tokens is predicted sequentially, conditioned on the previously generated tokens. This iterative token prediction process continues until the entire video is generated. See Supplementary \S\ref{apx:autogressive} for details. In \Cref{exp:video}, we show that even with a 1-step process, our method outperforms UniPi on the more challenging Cup Arrangement task, while using 8 steps further improves performance.

\begin{table}[t]
\small
  \centering
  \setlength{\tabcolsep}{0.25em}
  \scalebox{1}{
  \begin{tabular}{l|cc}
    \toprule
    & \bf Libero10 $\downarrow$ & \bf CupArrange $\downarrow$ \\
    \midrule
    \bf UniPi \cite{du2024learning} & 56.55   & 71.37 \\
    \bf \ours (1 step)  & 89.36 & 51.34 \\
    \bf \ours (8 steps) & \bf 51.10 & \bf 29.72 \\
    \bottomrule
  \end{tabular}
  }
  \caption{\small 
         \textbf{Video Generation Results.} Our method outperforms UniPi in both simulated (Libero10) and real-world (Cup Arrangement) environments. The masked autoencoder training enables autoregressive video generation, with 8 steps performing better than a single step. FVD \cite{unterthiner2018towards} results are reported.
         }
\label{exp:video}
\end{table}



\begin{figure*}[h!]
\begin{center}
\includegraphics[width=0.99\linewidth]{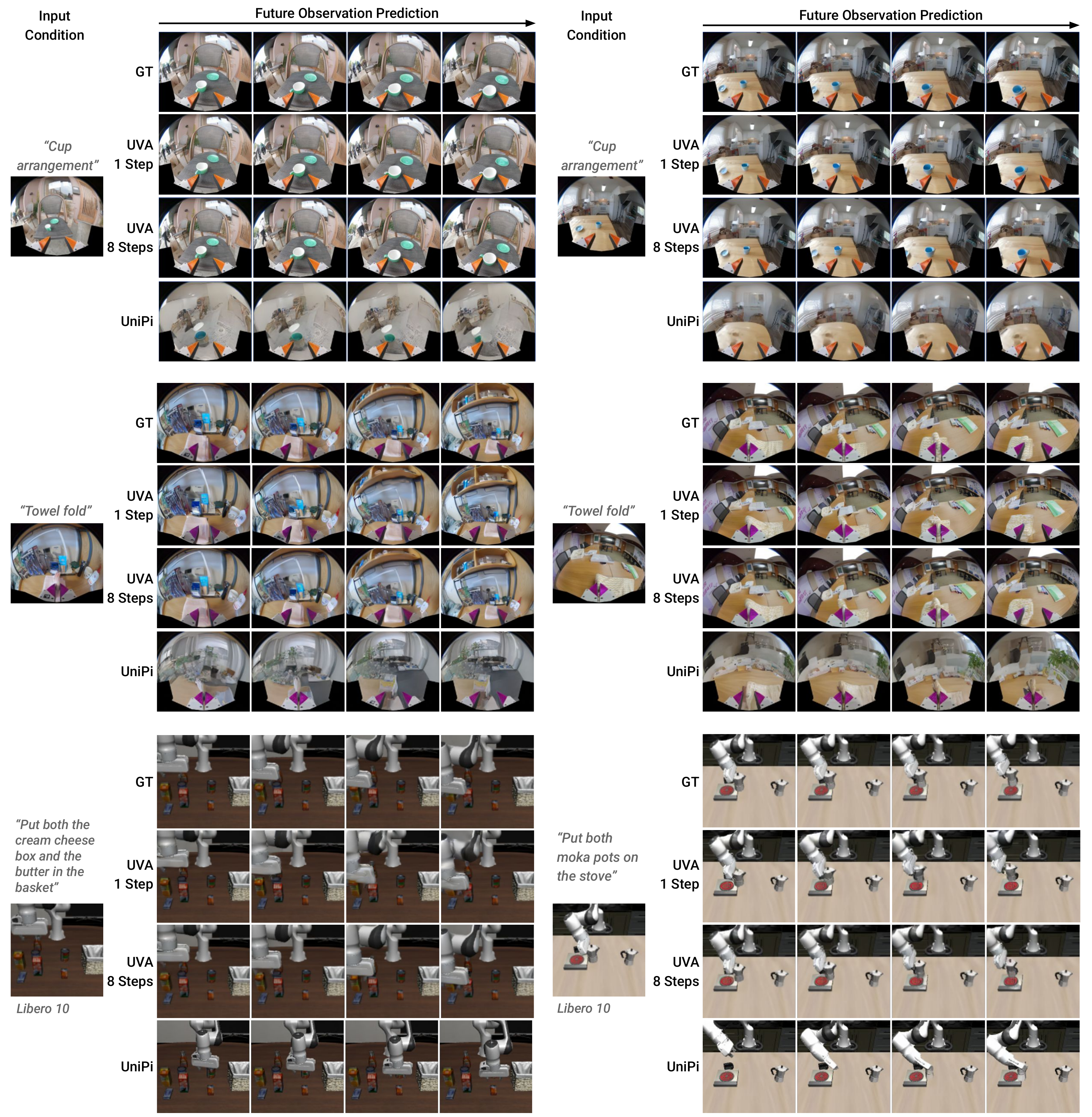}
\end{center}
\vspace{-10pt}
\caption{\small{\textbf{Video Generation Results on Validation Set.} UVA generates high-quality videos that closely match the ground truth, with 8 autoregressive steps further enhancing detail compared to a single autoregressive step. In contrast, UniPi occasionally produces blurry (Cup arrangement) or mismatched images (Towel fold and Libero10 left) and may fail to generate some objects (Libero10 right, the second moka pot). With the similar computational resources, UVA consistently outperforms UniPi.}}

\label{fig:video_gen_res2}
\vspace{-10pt}
\end{figure*}

\Cref{fig:video_gen_res2} shows the generated videos.
Conditioned on the history observations, UVA can predict future observations that closely match the ground truth. Even with a single autoregressive step, it can generate realistic video frames, and with additional steps, the details become more refined. In contrast, UniPi occasionally produces blurry images, such as the Cup Arrangement, or mismatched images, as seen in the Towel Folding task and the Libero10 task (left). Additionally, UniPi may fail to generate some objects entirely, as demonstrated in the Libero10 task (the second moka pot in the right figure). We believe that with more computational resources and larger video generation models, both UVA and UniPi could achieve further improvements. However, given similar computational resources, UVA consistently outperforms UniPi.



\begin{figure}[t]
    \centering    \includegraphics[width=0.99\linewidth]{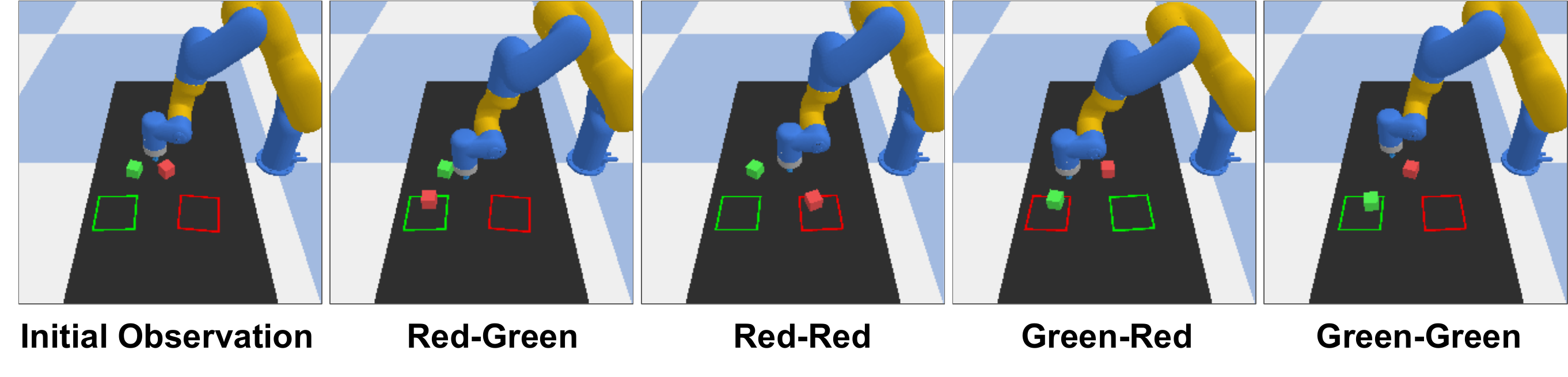}
    \vspace{-15pt}
    \caption{\small{\textbf{Forward Dynamics Model on Block Pushing Task.} During training, the robot pushes two blocks randomly to any target. During testing, the generated future image from \ours\ is used to select the proper action that moves a specific object to a specific target.}}
    \label{fig:forward_result}
    
    \vspace{3mm}
    
    \setlength{\tabcolsep}{0.3em}
    \scalebox{0.9}{
        \begin{tabular}{l|ccccc}
            \toprule
            & \bf R-R $\uparrow$ & \bf R-G $\uparrow$ & \bf G-R $\uparrow$ & \bf G-G $\uparrow$ & \bf Avg. $\uparrow$ \\
            \midrule
            \bf DP-C & 0.20 & 0.50 & 0.60 & 0.20 & 0.38 \\
            \bf \ours & \textbf{0.80 }& \textbf{0.70} & \textbf{0.50 }& \textbf{0.40} & \textbf{0.60} \\
            \midrule
            \bf GT-Dynamics & 0.80 & 0.80 & 0.70 & 0.70 & 0.75 \\
            \bottomrule
        \end{tabular}
    }
    \captionsetup{type=table}
    \caption{\small{\textbf{Success Rate on Block Pushing.} Our model functions as a forward dynamics model to guide the behavior of pretrained policy models, such as the DP-C \cite{chi2023diffusion}. DP-C alone achieves an 38\% success rate, while incorporating our model to generate future observations for trajectory selection increases the success rate to 60\%. Using a ground-truth simulator provides an upper bound success rate of 75\%.}}
    \label{exp:table_mpc}
\vspace{-10pt}
\end{figure}


\section{\ours~as a Forward Dynamics Model}
\label{sec:forward}
Our model can perform forward dynamics predictions \(\mathbf{O}_{t+1} = f_{\text{forward}}(\mathbf{O}_t, \mathbf{A}_{t})\). To evaluate its effectiveness, we use it to guide the behavior of a pretrained policy model, such as the DP-C. We evaluate this approach in a block-pushing environment, where the model is trained to push one block to a specified square and another block to a second square, each randomly assigned. During testing, we aim to control the policy to complete specific tasks, such as pushing the red block to the red square (\textit{R-R}) or the red block to the green square (\textit{R-G}). The evaluation considers four distinct settings, as illustrated in \Cref{fig:forward_result}. 
Each setting is tested 10 times with varying initial positions of the objects and the robot.
Notably, a perfect policy model in this setup would achieve a maximum average success rate of 50\%, since training only requires one block to be pushed to any square, while testing specifies exact target assignments for each block.

At each step, we sample 100 trajectories of 16 future actions using DP-C. The sampled actions, along with historical observations, are input into our model, which predicts future observations by functioning as a forward dynamics model. For each trajectory, we calculate a reward based on the predicted observations, selecting the trajectory with the highest reward. The reward is computed as the distance between the blocks and their target squares, which are identified from the predicted frames. We then execute the first 6 steps of the selected trajectory and resample new trajectories until the task is completed or the episode ends.
DP-C can complete the tasks with a success rate of 38\% in Table \ref{exp:table_mpc}. 
By leveraging our model to guide the policy, the success rate increases to 60\%.

For comparison, we evaluate the performance of using a ground-truth simulator to render the sampled trajectories and select the best ones. Even with the ground-truth simulator, the success rate is limited to 75\% due to suboptimal sampled trajectories and errors in object detection. While our model performs worse than the simulator, it still significantly enhances performance and guides the pretrained policy models to complete required tasks.

\section{\ours~as a Inverse Dynamic Model}
\label{sec:inverse}

In this section, we evaluate the effectiveness of the proposed method in an inverse dynamics setting, where actions are inferred as \(\mathbf{A}_t = f_{\text{inverse}}(\mathbf{O}_t, \mathbf{O}_{t+1})\) using UMI data.
For the UMI Cup Arrangement data, the robot's actions are naturally aligned with camera movements, enabling straightforward evaluation of action prediction accuracy using the ground truth camera poses obtained from motion capture (Mocap). Notably, our model had never encountered this test data during training.


\smallskip \noindent \textbf{Baselines:} 
As a comparison, we evaluate the actions (i.e., camera pose) generated by the inverse dynamics model used in UniPi \cite{du2024learning} and a well-engineered SLAM system used in UMI \cite{campos2021orb}, where the SLAM system requires an additional mapping. The UMI actions in the training data are derived from SLAM.

\begin{table}[t]
\small

  %
  \centering
  \begin{tabular}{lcc}
    \toprule
    & \bf {Position} $\downarrow$ & \bf {Rotation} $\downarrow$ \\
    \midrule
    \bf UniPi Inverse Dynamics & 1.92 cm & 2.21\textdegree \\
    \bf \ours~ & \textbf{0.75} cm & \textbf{1.11}\textdegree  \\
    \midrule
    \bf Visual Inertial SLAM & 0.41 cm & 0.30\textdegree \\
    \bottomrule
  \end{tabular}
  \caption{\small{\textbf{Inverse Dynamics.} L2 distances between predicted actions and ground truth actions from Mocap. The table compares positions and rotation results of UVA, the UniPi inverse dynamics model, and a well-engineered SLAM system \cite{campos2021orb}. UVA outperforms UniPi inverse dynamics model, while SLAM achieves the best accuracy but is more complex to implement.}}
\label{tab:mocap}
\vspace{-5pt}
\end{table}

\smallskip \noindent \textbf{Results:} For the actions predicted by each method, we compute the L2 distance to the ground truth actions from Mocap, as shown in \Cref{tab:mocap}.
The UniPi inverse dynamics model predicts actions from two consecutive images, which may lead to discontinuities in the predicted actions over time. In contrast, our method predicts 16 actions simultaneously, resulting in more consistent and temporally coherent predictions.
The action errors produced by SLAM were 0.41 cm for position and 0.30 degrees for rotation. 
While \ours~ exhibited slightly higher errors than SLAM, it still demonstrated strong performance with position errors under 1 cm and rotation errors around 1 degree. 
These results highlight the generalization capability of \ours\ for action prediction, even on unseen data, and suggest that it could serve as a viable alternative to SLAM, which is difficult to calibrate and suffers from a high failure rate. 







\section{Discussion}
\label{sec:conclusion}

\smallskip \noindent \textbf{Summary:} We propose a unified video-action model that jointly models and separately decodes video and actions. This design enables us to fully leverage video data as additional supervision, resulting in stronger performance and fast action prediction by skipping video decoding during inference. 
The framework inherently supports masking training, allowing it to fulfill various robotics functions, including acting as a policy, video model, forward and inverse dynamics model, and a combined policy and planner.
By fully utilizing video and action data in diverse configurations, our model reduces overfitting to specific tasks, outperforms previous methods, and demonstrates versatility for multi-purpose applications.

\smallskip \noindent \textbf{Limitation and Future Work:}
One limitation of our framework is that it does not currently leverage large amounts of actionless video data, which could provide valuable additional supervision. As a result, our method occasionally achieves only comparable performance to the DP-UMI on real-world tasks. We believe that pretraining the model on web-scale video datasets could significantly enhance its generalization capabilities, and we leave this exploration for future work.
Furthermore, our model can be naturally extended to predict modalities beyond video and action, such as sound and force, by incorporating additional diffusion heads, offering a more comprehensive and versatile framework. This remains a promising direction for future research.

\section*{Acknowledgment}
We would like to thank Huy Ha for his valuable advice on video recording and website design. 
We also thank Amber Xie, Austin Patel, Jennifer Grannen, Vincent de Bakker, John So, Max Du, and Vidhi Jain for their important feedbacks on the paper draft. 
We are grateful to Mengda Xu, Suneel Belkhale, Xiaomeng Xu, Fanqi Lin, Lirui Wang, and Tianhong Li for helpful discussions.
We would like to express our gratitude to Chi Cheng, Zhenjia Xu, Chuer Pan, Zeyi Liu, Huy Ha, Fanqi Lin, Yingdong Hu, and Zhaxizhuoma for their contributions to the shared UMI dataset.
Finally, we want to thank everyone who contributed their computing resources to help us train the models.

This work was supported in part by the Toyota Research Institute, NSF Award \#1941722, \#2143601, \#2037101, \#2132519, ONR Project \#N00014-22-1-2293 and the DARPA TIMAT project. The views and conclusions contained herein are those of the authors and should not be interpreted as necessarily representing the official policies, either expressed or implied, of the sponsors.

\bibliographystyle{plainnat}
\bibliography{references.bib}

\clearpage

\section{Supplementary Materials}
\label{sec:supplementary}
In this section, we first introduce the autoregressive video generation process in \S\ref{apx:autogressive} and then show more details of the simulation benchmarks (\S\ref{apx:simulation_benchmark}) and real-world benchmarks (\S\ref{apx:realworld_benchmark}) we used for evaluation. We finally report the inference speed details in \S\ref{apx:inference_speed}.

\subsection{Autogregressive Video Generation}
\label{apx:autogressive}
Our autoregressive video generation is based on the methods from \cite{chang2022maskgit} and \cite{li2024autoregressive}, originally designed for image generation, which we have extended to video generation.
In \cite{chang2022maskgit}, images are first converted into discrete visual codes using VQGAN \cite{esser2021taming}. During training, a subset of these visual codes is randomly masked and the model is trained to reconstruct them. During inference, the entire image is generated from an empty mask. This masked training approach enables the model to generate images autoregressively. The number of autoregressive steps can be adjusted during inference, with more steps leading to improved performance, as demonstrated in their paper. However, the visual discretization in VQGAN and training with discrete visual codes often leads to information loss, resulting in lower-quality images.
Li et al.~\cite{li2024autoregressive} address this limitation by using continuous latent representations instead of discrete code for image tokens. Their approach models each visual token's probability using a diffusion model, eliminating the need for vector quantization. This method has shown improved performance over previous approaches.

\begin{figure}[t]
\begin{center}
\includegraphics[width=1\linewidth]{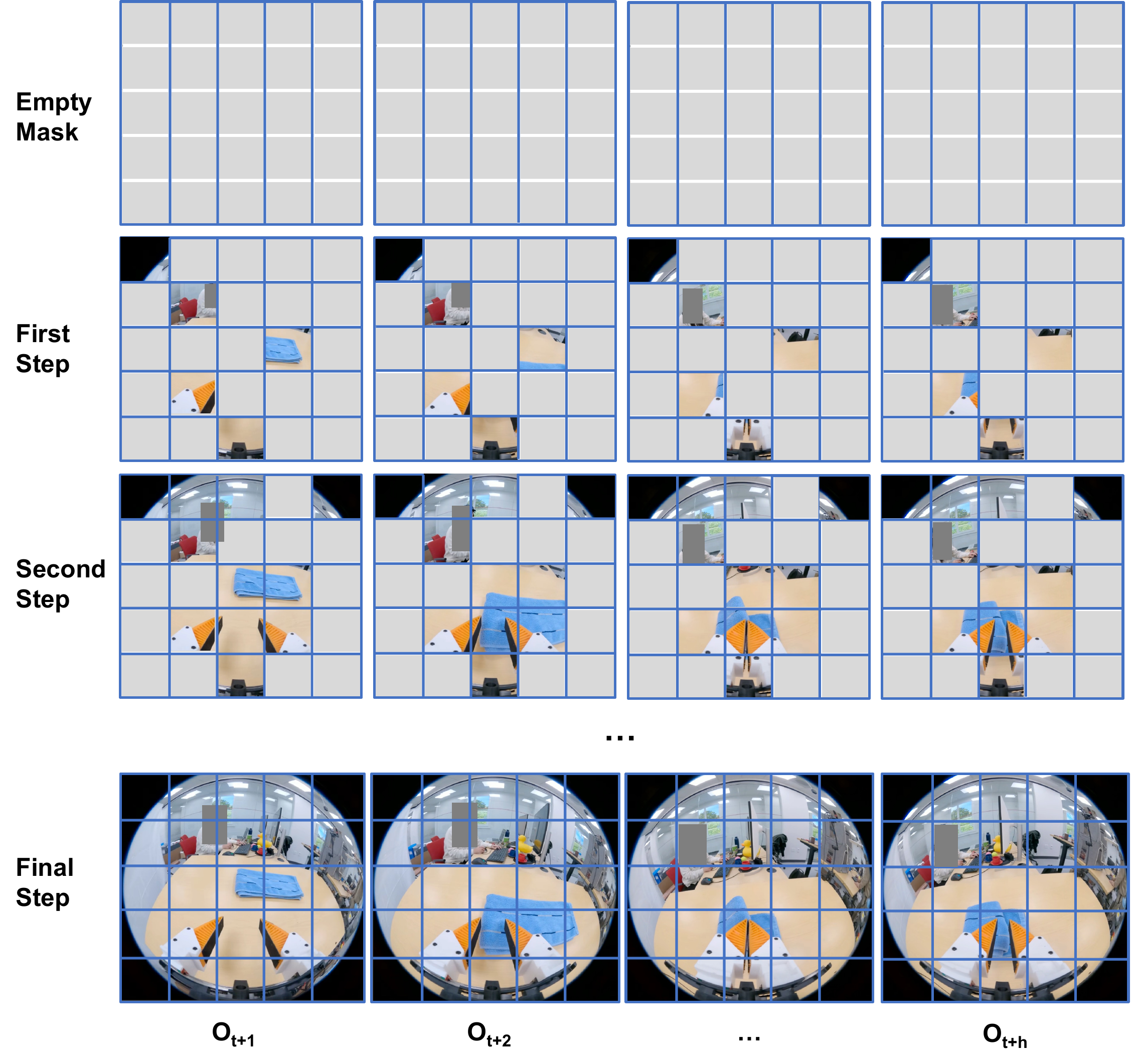}
\end{center}
\vspace{-10pt}
\caption{\small{\textbf{Autoregressive Video Generation.} Our method generates a video starting from an empty mask. Given a specified number of generation steps, the model produces a set of tokens at each autoregressive step, progressively constructing the video. If the generation step is set to 1, the entire video is generated in a single pass.}}
\label{apx_fig:autoregressive}
\end{figure}

Our method is similar to \cite{li2024autoregressive} in that it predicts continuous latent representations. 
These representations are then used as conditions of the diffusion heads to decode actions and video observations. The autoregressive generation process is shown in \Cref{apx_fig:autoregressive}.
If the autoregressive step is set to 1, the entire video is generated in a single pass.
Otherwise, with a predefined number of steps, the method generates the video autoregressively, completing the process in the specified number of steps.

Our model builds on the pretrained model (MAR-B) released by \cite{li2024autoregressive} but has undergone substantial modifications for the joint video and action modeling. 
Please check out our \hyperlink{https://github.com/ShuangLI59/unified_video_action}{code} for details.

\subsection{Simulation Benchmarks}
\label{apx:simulation_benchmark}
\vspace{2mm} \noindent
\textbf{Single-Task Evaluation:} For this setting, different policies are trained for different tasks. 
\begin{itemize}
    \item {PushT \cite{chi2023diffusion,florence2021implicit}}: requires the agent to push the gray ``T'' to align it with the target ``T'' at the center of the scene. The average success rate over 50 rollouts is reported in the main paper. 

    \item {Toolhang \cite{robomimic2021}}: requires a robot to insert a hook into a base and then hang a wrench.  It is one of the most challenging tasks in RoboMimic. We evaluate the average success rate over 50 environments with different random seeds.
    
\end{itemize}

\vspace{1mm} \noindent\textbf{Multi-Task Evaluation:} 
In multi-task evaluation, we train one policy for multiple task goals defined by image or text. 
\begin{itemize}
    \item {PushT-M}: We extend the \textit{PushT} task to include varying target ``T'' positions. This setting adds additional uncertainty and presents a greater challenge, requiring the agent to dynamically plan its trajectory and adapt to diverse spatial configurations. We reported the average reward on 50 environments with different random seeds. 

    \item {Libero10} \cite{liu2024libero}: is the most challenging task in the Libero benchmark due to its long-horizon requirements. This set comprises 10 tasks, each accompanied by a language goal description, such as ``put the red mug on the left plate and open the bottom drawer of the cabinet''. We use the same language encoder, \textit{e.g.}, CLIP, for all methods. Objects are initialized in varying locations within the scene.     
    Each task is evaluated over 10 runs with different random seeds, and we report the average reward across all 10 tasks, resulting in a total of 500 test episodes.
    
    
\end{itemize}

\subsection{Real-world Benchmarks}
\label{apx:realworld_benchmark}
\smallskip \noindent \textbf{Training Data:} We assess real-world performance using two publicly available benchmarks introduced by \cite{chi2024universal} and \cite{lin2024data}. Both benchmarks collect data using the handheld UMI \cite{chi2024universal} device. Thanks to the UMI's design, these datasets can be directly utilized to train various models and evaluate them on our robotic setup, which consists of a single ARX X5 robotic arm. We used three tasks for training: 
\begin{itemize}
    \item  Cup Arrangement \cite{chi2024universal}: requires the robot to first rotate the cup with the handle oriented to the left of the robot, then pick up an espresso cup and place it onto a saucer. Success is achieved when the cup is properly placed on the saucer and the handle lies within \( \pm 15^\circ \) of the exact left alignment. 
    \item Towel Folding \cite{lin2024data}: requires the robot to grasp the left edge of the towel and move it to the right, folding it in half. 
    \item Mouse Arrangement \cite{lin2024data}: requires the robot to pick up the mouse and place it on the mouse pad. 
\end{itemize}

\vspace{1mm} \noindent
\textbf{Single-Task Evaluation:} We train a single-task policy on the Cup dataset and directly compare it to the DP-UMI model provided by the authors \cite{chi2024universal}. Both methods are trained using the same dataset.
The cup and saucer are initially placed in varying positions and orientations. Across 20 test rollouts, the robot needs to rotate the cup before placing it on the saucer in 85\% of the cases. Additionally, we include challenging scenarios where the cup is positioned at the boundary of the robot arm’s reach.
We evaluate each method across 20 rollouts with different initial configurations. The average success rate is reported in the main paper.

\vspace{1mm} \noindent
\textbf{Multi-Task Evaluation:} We train one model with all three tasks and then evaluate the policy performance on each task independently. 
The three public datasets contain a total of 6,764 episodes. We randomly selected 500 episodes from each dataset and combined them into a dataset with 1500 episodes to train both our model and DP-UMI \cite{chi2024universal}.
To ensure a wide testing distribution, we evaluate 20 different scenarios, including six initial configurations of objects and the robot, six different distractor objects, five distinct backgrounds, and an unseen gripper type (training is conducted with only orange and purple grippers, while testing includes a green gripper).

    

\subsection{Inference Speed Measurement and Decomposition}
\label{apx:inference_speed}

For the simulation results in \Cref{exp:policy_results_sim}, we measure the inference speed using a server with NVIDIA L40 GPUs for all of the models. For the real-world results in \Cref{exp:policy_results_real}, we measure the inference speed on an NVIDIA RTX 3080 GPU to match real-world deployment scenarios.

We also decompose the inference time of each component in our model under the real-world deployment scenario:

\begin{table}[h]
\small

  %
  \centering
\begin{tabular}{lc}
    \toprule
    \bf {Modules / Tasks} & \bf {Inference time (ms)} $\downarrow$ \\
    \midrule
    \bf VAE Image Encoder & 40 \\
    \bf Transformer (Attention) & 40 \\
    \bf Transformer (Flash Attention) & 30 \\
    \bf Action Diffusion (16 steps)& 15 \\
    \bf Action Diffusion (100 steps)& 93 \\
    \bf Rest of the model & $ < 1$ \\
    \bf Video Diffusion (16 steps) & 100 \\
    \bf Video Diffusion (100 steps) & 625 \\
    \midrule
    \bf \ours\ (16 steps) & 95 \\
    \bf \ours\ (16 steps Flash Attention) & 85 \\
    \bf \ours\ (100 steps) & 173 \\
    \bf \ours\ (100 steps Flash Attention) & 163 \\
    \bf Policy+Planner (16 steps) & 195 \\
    \bf Policy+Planner (100 steps) & 798 \\
    \bottomrule
\end{tabular}
  \caption{\small{\textbf{Inference Time Decomposition.} We measure the inference time for each module and the total runtime of \ours\ as a policy model, using 16 policy steps and 100 diffusion steps.
  }}
\label{tab:speed}
\end{table}

The pretrained VAE image encoder requires 40ms, while the transformer module takes approximately 40ms, which can be reduced to 30ms using flash attention. The action prediction head takes 15ms with 16 diffusion steps and 93ms with 100 diffusion steps. In our policy learning experiment in \Cref{exp:policy_results_real}, we chose 16 diffusion steps. However, using 100 diffusion steps results in smoother action predictions and a higher success rate, though at the cost of slower performance.
In total, \ours\ requires 95ms to predict an action trajectory with 16 actions when using 16 diffusion steps, and 173ms when using 100 diffusion steps. Incorporating flash attention can reduce 10ms for both.

We also reported the inference speed for video generation, even though video generation was skipped during policy inference. Video generation involves a higher latent dimensionality and requires more denoising steps. Performing 16 video diffusion steps takes 100ms, while 100 steps take 625ms. Therefore, skipping video generation significantly speeds up policy inference.

\subsection{Utilize Action-Free Video} 
UVA can leverage action-free video data by (1) pretraining on video generation tasks and (2) co-training with video-action data to learn a joint latent representation. We conducted an experiment where UVA is first pretrained for video generation using the Human Video dataset from~\cite{clark2025action}, which contains 3,175 human-only videos. We then finetune UVA on the combined Human Video and LIBERO-10 datasets using the Masked Training strategy. With the action-free human videos, the accuracy can be further improved as shown in \Cref{tab:humandata}.

\begin{table}[h]
\small
  \centering
\begin{tabular}{lcc}
    \toprule
    \bf {Model} & \bf {30 test} $\uparrow$ & \bf {500 test} $\uparrow$ \\
    \midrule
    \bf UVA & 0.93 & 0.90 \\
    \bf UVA + Human Data & 0.97 & 0.91 \\
    \bottomrule
\end{tabular}
  \caption{\small{\textbf{UVA with additional human data further improves performance.} Results on Libero10 across 10 tasks. In the 30-test setting, each task is evaluated 3 times with different random seeds. In the 500-test setting, each task is evaluated 50 times.}}
\label{tab:humandata}
\end{table}







\begin{table*}[t]
\centering
\scriptsize
\resizebox{0.75\linewidth}{!}{

\begin{tabular}{l|ccc|ccc}
    \toprule
    \textbf{Masking Strategy} & \multicolumn{3}{c|}{\textbf{Application-dependent}} & \multicolumn{3}{c}{\textbf{Application-independent}} \\
    \textbf{Mask Ratio} & 25\% & 50\% & 75\% & 25\% & 50\% & 75\% \\
    \midrule
    Policy (Success) $\uparrow$ & \bf 0.84  &  \bf 0.84 & 0.74 &  0.86 & \bf 0.87 & 0.84 \\
    Video Generation (FVD) $\downarrow$ & \bf 127.09 & 161.72 & 166.45 & \bf 181.68 & 200.08 &  262.93 \\
    Forward Dynamics (FVD) $\downarrow$ & \bf 129.77 & 167.10 & 167.40 & \bf 174.85 & 203.59 &  264.96\\
    Inverse Dynamics (L2) $\downarrow$ & \bf 17.54 & 25.98 & 47.79 & 14.14 & \bf 13.55 & 14.10 \\
    \bottomrule
\end{tabular}

\small
\vspace{-3pt}



}
\caption{Evaluation of different masking strategies (application-dependent vs. application-independent) across various mask ratios.}
\label{tab:mask_strategy_comparison}
\end{table*}

\subsection{Impact of masking strategy.} 
In paper \Cref{fig:overview}, our masked pattern is based on the input-output structure of each application.
We added two additional masking methods on PushT-M:
1) Application-dependent: The masking follows the input-output structure of each application, e.g., video generation uses history observations to predict future ones. We thus randomly mask the history observations.
2) Application-independent: randomly mask the inputs, regardless of task semantics.
The results are reported in \Cref{tab:mask_strategy_comparison}
%

Policy learning and video generation are evaluated by success rate and FVD. 
Forward dynamics is evaluated by FVD on videos generated conditioned on actions.
Inverse dynamics is evaluated by L2 error.
Overall, in the ``application-dependent'' setting, higher mask ratios degrade overall performance.
In the ``application-independent'' setting, 25\% masking works best for video generation and forward dynamics, while 50\% is optimal for policy and inverse dynamics.


\end{document}